\documentclass[runningheads]{llncs}
\usepackage[T1]{fontenc}
\usepackage{graphicx}
\usepackage{amsmath}
\usepackage{multirow}
\usepackage{booktabs}
\usepackage{amssymb}
\usepackage[colorlinks=true,citecolor=blue,urlcolor=blue]{hyperref}
\usepackage[misc]{ifsym} 
\usepackage{subfigure}
\usepackage{color}
\usepackage{algorithm}
\usepackage{algorithmic}
\begin{document}
	\title{FedMLP: Federated Multi-Label Medical Image Classification under Task Heterogeneity}
	\titlerunning{FedMLP}
	%

      \author{Zhaobin Sun\thanks{Equal contribution.}\inst{1}
		 	\and
		     Nannan Wu\inst{\star 1} 
		 	\and
		 	Junjie Shi\inst{1} 
		 	\and
		 	Li Yu\inst{1} 
		 	\and
		 	Xin Yang\inst{1} 
		 	\and 
		 	\\Kwang-Ting Cheng\inst{2} 
		 	\and 
		 	Zengqiang Yan\textsuperscript{1(\Letter)} 
		 }
	
	 \authorrunning{Z. Sun et al.}
	
	 \institute{School of Electronic Information and Communications, Huazhong University of Science and Technology \\
		 	\email{\{zbsun,wnn2000,shijunjie,hustlyu,xinyang2014,z\_yan\}@hust.edu.cn} 
		 	\and School of Engineering, Hong Kong University of Science and Technology 
		 	\email{timcheng@ust.hk}
		 }
	\maketitle              
        \setcounter{footnote}{0}
	\begin{abstract}
		
		Cross-silo federated learning (FL) enables decentralized organizations to collaboratively train models while preserving data privacy and has made significant progress in medical image classification. One common assumption is task homogeneity where each client has access to all classes during training. However, in clinical practice, given a multi-label classification task, constrained by the level of medical knowledge and the prevalence of diseases, each institution may diagnose only partial categories, resulting in task heterogeneity. How to pursue effective multi-label medical image classification under task heterogeneity is under-explored. In this paper, we first formulate such a realistic label missing setting in the multi-label FL domain and propose a two-stage method FedMLP to combat class missing from two aspects: pseudo label tagging and global knowledge learning. The former utilizes a warmed-up model to generate class prototypes and select samples with high confidence to supplement missing labels, while the latter uses a global model as a teacher for consistency regularization to prevent forgetting missing class knowledge. Experiments on two publicly-available medical datasets validate the superiority of FedMLP against the state-of-the-art both federated semi-supervised and noisy label learning approaches under task heterogeneity. Code is available at \url{https://github.com/szbonaldo/FedMLP}.
		
	\end{abstract}
	\section{Introduction}
	
\begin{figure}[!t]
		\centering
		\includegraphics[width=0.6\textwidth]{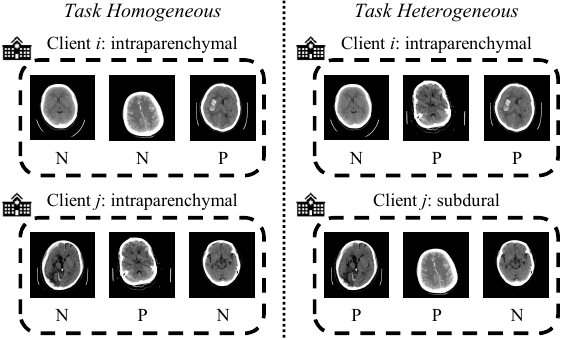}
		\caption{Illustration of two decentralized data settings.}
		\label{fig:bg}
	\end{figure}

	Attributed to heightened privacy concerns, merging multiple medical image datasets into a unified one is often prohibited, presenting additional challenges in developing deep neural models for automated disease classification. Recently, federated learning (FL) \cite{mcmahan2017communication}, a technique that enables collaborative model training across decentralized sources without compromising privacy, has demonstrated significant potential in overcoming this obstacle \cite{wu2023federated,jiang2022dynamic,liu2021federated,FLSurveyandBenchmarkforGenRobFair_arXiv23,FCCLPlus_TPAMI23,FedA3I,FedISM}. Nevertheless, current research predominantly operates under a critical assumption: cross-client data\footnote{Data includes images and corresponding labels.} is \textit{\textbf{task homogeneous}} \cite{FCCL_CVPR22}, as shown in Fig. \ref{fig:bg}. However, in practical data acquisition, a prevalent encountered scenario involves diverse institutions amassing data that is \textit{\textbf{task heterogeneous}}, depending on their particular areas of interest. As depicted in Fig. \ref{fig:bg}, one hospital may prioritize intraparenchymal hemorrhage and thereby assemble a dataset. Conversely, another institution might gather data specifically for subdural hemorrhage reflecting its distinct focal points. 
	This diversity in class interests and tasks among various clients introduces extra complexity to implementing FL, as the knowledge encapsulated by available labels is not universally shared across participants.
	
	In this study, we conceptualize each client's task as a multi-label classification problem characterized by partial annotations, reflecting that the areas of interest for certain clients represent a subset of the collective focus areas across all clients. Against this backdrop, our research is directed at tackling the critical challenge: \textit{\textbf{how to achieve data-efficient federated multi-label learning (FMLL) against heterogeneity in partially labeled classes}}? Given diverse imperfect data caused by this heterogeneity, approaches in federated noisy label learning (FNLL) \cite{wu2023fednoro,xu2022fedcorr,jiang2022towards,wu2023learning} and federated semi-supervised learning (FSSL) \cite{li2023class,liang2022rscfed,liu2021federated,cho2023local} fall short of effectively addressing it. Their designs, predominantly aimed at handling imperfect data in a homogeneous manner and focusing on multi-class classification, are sub-optimal when applied to FMLL scenarios. Thus, navigating this challenge remains a relatively untapped area of exploration.
	
	To tackle this problem, the crux is to transfer knowledge from diverse tasks to each local client. We propose a prototype-based, two-stage method, named \textbf{Fed}erated \textbf{M}ulti-\textbf{L}abel learning with \textbf{P}artial annotation (\textbf{FedMLP}). The initial stage involves a model warm-up to establish high-quality class prototypes. It is trained by a multi-label Weighted-Partial-Class (WPC) loss, combined with logit adjustment \cite{menon2020long}, to minimize the misclassification of false negatives in negative classes and mitigate the effects of class imbalance. Following the creation of initial class prototypes, the subsequent stage employs a dynamic Self-adaptive Threshold (ST) to select pseudo-labeling classes of samples whose features are more similar to class-wise prototypes, thus progressively facilitating task-specific knowledge transfer to each client, guided by further training under pseudo labels. To address the challenge of classes lacking pseudo labels, Consistency Regularization (CR) between global and local models is employed to diminish the potential negative effects of inaccurate knowledge within these classes.
	
	The main contributions are summarized as follows. (1) An rarely-explored FMLL setting considering task heterogeneity. (2) A novel solution FedMLP to address heterogeneity in partially labeled classes. (3) Superior performance on two real-world medical datasets against SOTA FSSL and FNLL methods.

	\section{Methodology}
	\subsection{Preliminaries and Overview}
	
\begin{figure}[!t]
		\centering
		\includegraphics[width=0.95\textwidth]{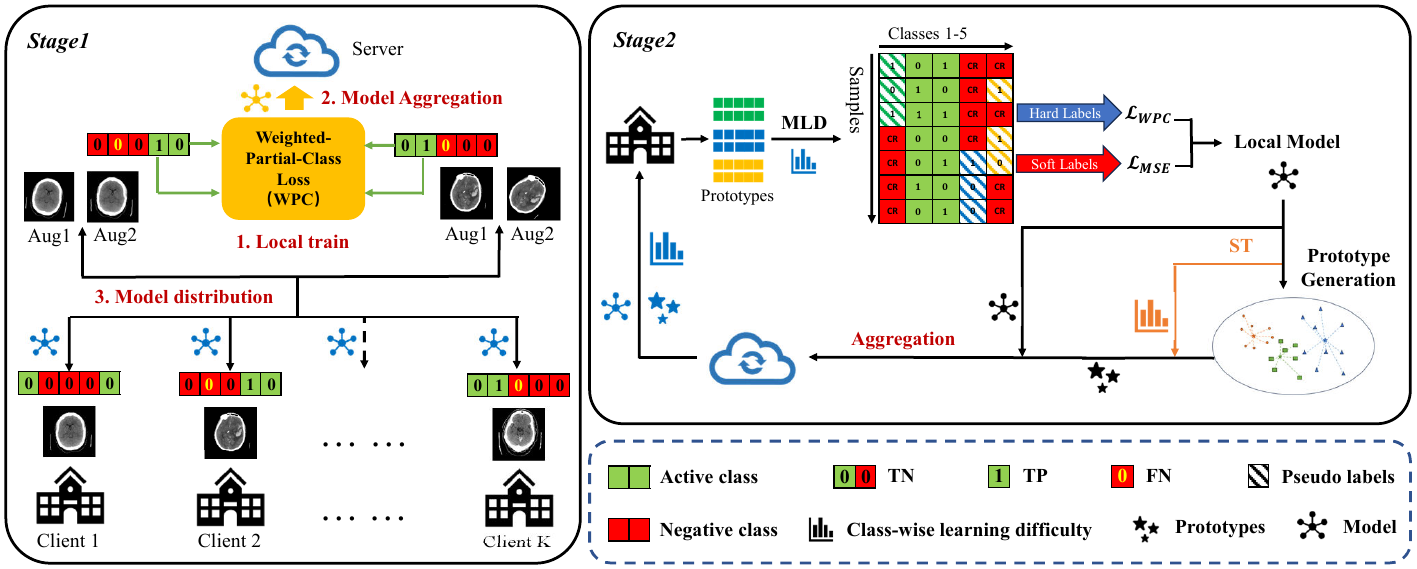}
		\caption{Overview of FedMLP. Colors red and green represent missing and labeled classes respectively.}
		\label{overview}
	\end{figure}

	Given $K$ participants, any $k$-th participant holds a dataset $D_k = \{(x_i, y_i)\}_{i=1}^{N_k}$ with $N_k$ samples. Here, $(x_i, y_i)$ is an image-label pair, with $x_i \in \mathcal{X} \subseteq \mathbb{R}^d$ and $y_i \in \mathcal{Y} = \{0, 1\}^C$, where $C$ is the total number of classes. In scenarios with missing labels, each client $k$ might not have labels for certain classes. To describe this, we define an active class set $AC_k$ and a negative class set $NC_k$, with $AC_k$ containing indices of classes with labels and $NC_k$ comprising those without, ensuring $AC_k \cap NC_k = \emptyset$ and $AC_k \cup NC_k = [C]$. In the following context, active classes and negative classes are equivalent to the categories of labeled and missing classes respectively, and can be used interchangeably. Without data privacy leakage, we can get information of client-wise class-level annotation distribution $\mathbb{S} = \{\mathbb{S}_i\}_{i=1}^{C}$ at the server side, where $\mathbb{S}_i$ denotes the set of client indexes recognizing class $i$, $\mathbb{S}_i \neq \emptyset$. In our setting, due to diversity in class interest, $\left| \mathbb{S}_i \right|$ is not identical for each class. A larger $\left| \mathbb{S}_i \right|$ implies a more common disease $i$, which is easier to learn than other classes, denoted as "hot classes", otherwise as "cool classes".
	The objective is to develop a global model capable of effectively detecting both hot and cool classes. The overview of FedMLP is illustrated in Fig. \ref{overview}. 
	
	\subsection{Warm-up with Missing Labels}
	\label{warmup}
	Different from the typical FSSL scenario where at least one client or server has instances with complete labels, in FMLL with partial labels, from the perspective of each client, the only way to learn knowledge of local-unknown classes is from other clients. Previous works regard missing classes as negative \cite{bucak2011multi,chen2013fast,sun2010multi,wang2014binary} and use a binary cross entropy(BCE) loss on the multi-label setting formulated as
	\begin{equation} \label{oribce}
		L_{BCE}(x_i, y_i) = -\frac{1}{C} \sum_{j = 1}^{C} (y^j_i \cdot \log(\hat{y}^j_i)+
		(1-y^j_i) \cdot \log(1-\hat{y}^j_i)),
	\end{equation} 
	where $\hat{y}^j_i$ denotes the predicted probability of class $j$ by the Sigmoid function of sample $i$. Trained through this, clients tend to predict negatively if a label is missing, caused by the biased classifier and dense features, making the global model struggle to recognize cool classes completely. Though hot-class knowledge may be compensated by model aggregation, there is a risk of catastrophic forgetting when data distribution is quite heterogeneous.
	
	Considering the different prevalences of diseases, BCE loss may make the model biased towards updating the majority classes. Related works like Logit Adjustment (LA) \cite{menon2020long} were proposed for multi-class classification, fine-tuning model output by calculating the positive rates of each category and mitigating the impact of class imbalance, formulated as
	\begin{equation} \label{logit adjustment}
		l^\prime_c = l_c + \tau \cdot \log(\pi_c),
	\end{equation} 
	where $l_c$ denotes the logit of class $c$, $\tau$ is a hyper-parameter set as 1 by default, and $\pi_c$ is the sample proportion of class $c$ in the training set. In this work, we further extend LA to the multi-label setting:
	\begin{equation} \label{logit adjustment multi label}
		\hat{y}^{c\prime}_i = \frac{\hat{y}^c_i \cdot \pi_{c,1}}{\hat{y}^c_i \cdot \pi_{c,1}+(1-\hat{y}^c_i)\cdot \pi_{c,0}},
	\end{equation} 
	where $\pi_{c,1}$ represents the proportion of positive samples in class c, and the sum of $\pi_{c,1}$ and $\pi_{c,0}$ equals to 1.
	Considering partial annotation makes the local model lazy to predict missing classes as negative, thus we propose partial-class loss to exclude the disturbance of false-negative labels, forming a Weighted-Partial-Class loss
	\begin{equation} \label{weightpartialbce}
		\begin{split}
			L^k_{WPC}(x_i, y_i) = -\frac{1}{C} \sum_{j \in AC_k} ((1-y^j_i) \cdot \log(1-\hat{y}^{j\prime}_i) 
			+y^j_i \cdot \log(\hat{y}^{j\prime}_i)).
		\end{split}
	\end{equation} 
	At the first $t_1$ communication rounds, we take $L^k_{WPC}$ as the objection of local optimization and continue to use FedAvg for model aggregation. To enhance sample diversity, we further utilize data augment techniques, inspired by semi-supervise learning \cite{sohn2020fixmatch}. To this end, we get a warmed-up model to generate the representation $F^k_i$ of sample $x_i$ in client $k$ with acceptable quality. 
	
	\subsection{Missing Label Detection with Global Active Class Prototypes}
	\label{corr}
	Toward the phenomenon where traditional training methods tend to ignore cool classes under class interest diversity (\textit{see the supplementary material}), we use prototypes to detect and choose confident samples for missing classes (Stage 2), enabling each client to gradually learn the missing knowledge of local samples.
	
	\subsubsection{Prototype Generation.} 
	For each client $k$, we only consider local active classes $AC_k$ for prototype calculation to improve feature quality. Each class $c$ has dual prototypes: $P^{k,c}_0$ and $P^{k,c}_1$ denoting local prototypes of negative and positive for class $c$ in client $k$, updated at the end of each round after warm-up, defined as
	\begin{equation} \label{localproto0}
		P^{k,c}_n = \frac{\sum_{i=1}^{N_k} \mathbb{I}(y^c_i=n) \cdot F^k_i }{\left| D_{k, y^c_i=n} \right|}, n \in \{0,1\}
	\end{equation} 
	where $c \in AC_k$ and $\left| D_{k, y^c_i=1} \right|$ is the number of samples being positive in class $c$. Each client only delivers its local prototypes and updated model to the server, maintaining the same communication cost as FedAvg. Then, the server aggregates clients' models and prototypes and sends the global prototypes of all classes and the model to clients. Global active class prototypes are updated by
	\begin{equation} \label{globproto0}
		P^c_n = \frac{\sum_{i=1}^{K} \mathbb{I}(i \in \mathbb{S}_c) \cdot P^{i,c}_n }{\left| \mathbb{S}_c \right|}, n \in \{0, 1\}.
	\end{equation}  
    According to previous research\cite{FPL_CVPR23}, taking the average of multiple iterations of features can effectively reduce the privacy leakage.
	\subsubsection{Missing Label Detection (MLD).} 
	Typical FNLL clustered sample losses to recognize noisy clients and samples \cite{wu2023fednoro,xu2022fedcorr} and used pseudo labeling for correction. To some extent, our scenario can be considered a special case with an explicit noise transition matrix, which completely turns missing classes to negative. Utilizing global active class prototypes, we already have the confidence of any instance $i$ for a local negative class $c$ in client $k$
	\begin{equation} \label{confidence}
		Z^{k,c}_i = \textbf{cos}(P^c_0, F^k_i) - \textbf{cos}(P^c_1, F^k_i),
	\end{equation} 
	where $\textbf{cos}$ is cosine similarity. $Z^{k,c}_i < 0$ indicates the sample is more likely to be positive in class $c$, otherwise more likely to be negative. $\left| Z^{k,c}_i \right|$ approaches 0 for uncertain samples. Following this, we propose a novel criterion to select high-confidence samples for pseudo labeling, according to
	\begin{equation} \label{hardlabels0}
		\mathbb{P}^k_{c,0} = \{i \mid Z^{k,c}_i > \textbf{Rank}(\mathbb{Z}^{k,c}, \tau_0), Z^{k,c}_j \in \mathbb{Z}^{k,c} \geq 0 \},
	\end{equation}
	\begin{equation} \label{hardlabels1}
		\mathbb{P}^k_{c,1} = \{i \mid -Z^{k,c}_i > \textbf{Rank}(-\mathbb{Z}^{k,c}, \tau_1), Z^{k,c}_j \in \mathbb{Z}^{k,c} < 0 \},
	\end{equation}
	where $\textbf{Rank}(\mathbb{Z}^{k,c}, \tau)$ denotes the set of top $\tau$ percent elements $\mathbb{Z}^{k,c}$. If a sample is in $\mathbb{P}^k_{c,0}$ or $\mathbb{P}^k_{c,1}$, it will be permanently tagged with the label 0 or 1. In later rounds, $\mathbb{Z}^{k,c}$ only includes residual samples until all samples have pseudo labels. $\tau_0$ and $\tau_1$ denote the selection ratios of samples with pseudo labels to alleviate erroneous information. Given more difficult classes to learn, sample selection is more discreet indicating a smaller $\tau$. To automate this, we propose a self-adaptive ratio mechanism as discussed in the following.
	
	\subsubsection{Self-adaptive Threshold.} 
	The way to determine the class learning difficulty is crucial to adjusting the sample selection ratio automatically. In the local view, as the active classes are deemed to have correct labels, it may reduce the difficulty gap between classes. Thus, we evaluate negative class difficulty in each client using its local data and average them on the server side at the end of each communication round, via uncertainty estimation:
	\begin{equation} \label{difficulty}
		d^k_c = \frac{\sum_{i=1}^{N_k} \mathbb{I}(\hat{y}^c_i<L \cup \hat{y}^c_i>R)}{\left| D_k \right|},
		\text{ and } 
		d^G_c = \sum_{k\in \mathbb{S}_c} \frac{\left| D_k \right|}{\sum_{i\in \mathbb{S}_c} \left| D_i \right|} \cdot d^k_c,
	\end{equation}
	where $d^k_c$ and $d^G_c$ denote the learning degree of class $c$ in client $k$ and the global learning degree respectively, $L, R$ are hyper-parameters. When a disease is easy to recognize, the numerator of $d^k_c$ increases, implying more samples can be selected to train with pseudo labels. Then, $\tau_0$ and $\tau_1$ are formulated as $\tau_0 = d^G_c \cdot T_0$ and $\tau_1 = d^G_c \cdot T_1$ where hyper-parameters $T_0$ and $T_1$ controls the selected ratios of negative and positive samples.
	
	\subsection{Consistency Regularization of Uncertain Classes} \label{steady}
	The training process will become unstable if only partial classes are updated, \textit{e.g.}, knowledge quickly fades away when gradients are not computed for a minority category. Thus, we further utilize the global model as a teacher to avoid clients forgetting the knowledge about negative classes without pseudo labels. In this way, each sample will be trained by at most three labels including the correct active class label, the pseudo label, and the soft label prompted by the global model. We use the weighted-partial-class loss mentioned in Sec. \ref{warmup} for the first two labels and the mean-squared (MSE) loss for the soft labels.
	
	\section{Experiments}
	
	\subsection{Experimental Setup}
	
	\subsubsection{Dataset.} 
	
	Two multi-label medical image datasets are adopted.
	\textit{\textbf{ICH}}: The RSNA Intracranial Hemorrhage dataset \cite{flanders2020construction}, with over 800,000 CT images, is condensed to a subset of 180,000 instances by mixing all lesion samples and partial healthy samples. From this, 25,000 samples are further used for training, with the rest designated as the test set.
	\textit{\textbf{ChestXray14}}: The ChestXray14 dataset \cite{wang2017chestx} contains chest X-ray images from over 30,000 patients, of which PA-view samples are selected and split into training and testing sets in a 7:3 ratio. 
	Both datasets are pre-processed following protocols in \cite{liu2021federated,wu2023fednoro}, with an equal distribution of training data among clients.
	
	\subsubsection{Partial Label Generation.} 
	
	We randomly remove an equal number of categories from each client, ensuring that no client possesses a complete set of labels. \textit{ICH} comprises five subtypes, whereas \textit{ChestXray14} includes 14 categories. Our focus narrows to the eight categories with the highest incidence of positive cases. To validate the performance in dealing with extreme label scarcity where each client labels only one category, we adjust the number of clients to match the number of categories, resulting in five clients for \textit{ICH} and eight for \textit{ChestXray14}.
	
	{\renewcommand{\arraystretch}{1.1}
		\begin{table}[!t]
			\centering
			\caption{Quantitative comparison on \textit{ICH} under four levels of partial annotation settings. The results (\%) based on the last-epoch model on the testing set are reported. The best and second-best results are marked in bold and underlined.}\label{compare ich}
			\resizebox{\linewidth}{!}{
				\begin{tabular}{c|c|ccc|ccc|ccc|ccc|ccc}
					\toprule
					\hline
					\multirow{2}{*}{Type} & \multirow{2}{*}{Method}  &     \multicolumn{3}{c|}{Missing 1 class}  &\multicolumn{3}{c|}{Missing 2 classes}        & \multicolumn{3}{c|}{Missing 3 classes}    & \multicolumn{3}{c|}{Missing 4 classes}  & \multicolumn{3}{c}{Average}           
					\\ \cline{3-17} 
					&           & BACC       &AUC        &\multicolumn{1}{c|}{mAP}            
					& BACC       &AUC        &\multicolumn{1}{c|}{mAP}
					& BACC       &AUC        &\multicolumn{1}{c|}{mAP}
					& BACC       &AUC        &\multicolumn{1}{c|}{mAP}
					& BACC       &AUC        &mAP
					\\ \hline
					\multirow{2}{*}{Base} 
					&FedAvg         & 74.58     & 90.65     & 70.32          
					& 66.13     & 88.85     & 64.76      & 58.64     & 85.43     & 54.53      
					& 50.00     & 76.32     & 32.62
					& 62.34     & 85.31     & 55.56\\  \cline{2-17}
					& FedAvg-PL      & 80.52     & 91.32     & 70.94 & 76.12     & 87.88     & 58.98      & 63.40     & 83.12     & 44.18      
					& 61.46     & 79.48     & 40.60 
					& 70.38     & 85.45     & 53.68\\  \cline{1-17}
					\multirow{4}{*}{FSSL} 
					&RSCFed        & 80.26     & 90.83     & 70.79          
					& 79.09     & 89.24     & 67.27      & 67.55     & 85.40     & 61.43      
					& 67.16     & 71.16     & 36.89           
					& 73.52     & 84.16     & 59.10\\  \cline{2-17}
					&FedFixMatch       & 80.75     & \underline{91.52}     & \underline{71.03}          
					& 80.75     & 90.19     & 69.10      & 66.26     & 77.75     & 46.35      
					& 50.00     & 74.90     & 31.05           
					& 69.44     & 83.59     & 54.38\\  \cline{2-17}
					&FedIRM       & 79.77     & 90.93     & 70.63          
					& 72.78     & 87.12     & 63.17      & 66.10     & 83.92     & 49.68      
					& 69.39     & 77.19     & 36.73           
					& 72.01     & 84.79     & 55.05\\  \cline{2-17}
					& CBAFed      & 76.13     & 91.43      & 70.58
					& 71.97     & 87.50      & 63.27      & 55.66     & 74.85     & 43.94      
					& 52.28     & 72.15     & 34.99           
					& 64.01     & 81.48     & 53.20\\  \cline{1-17}
					\multirow{2}{*}{FNL} 
					&FedLSR       & 75.81     & 91.28     & 70.52          
					& 65.48     & 89.04     & 66.76      & 56.67     & 84.62     & 52.70      
					& 50.00     & 74.14     & 29.73           
					& 61.99     & 84.77     & 54.93\\  \cline{2-17}
					& FedNoRo      &\underline{82.47}      &90.92      &70.88  & \underline{81.90}     & \underline{91.12}     & \underline{69.53}      & \underline{80.98}     & \textbf{90.99}     & \textbf{65.73}      
					& \underline{74.64}     & \underline{85.09}     & \underline{59.28}       
					& \underline{80.00}     & \underline{89.53}     & \underline{66.36}\\  \cline{1-17}
					\multirow{1}{*}{Ours}
					& FedMLP       & \textbf{83.50}      &\textbf{92.84}      &\textbf{73.80}  & \textbf{82.26}     & \textbf{91.75}     & \textbf{70.52}      & \textbf{81.10}     & \underline{90.87}     & \underline{65.53}      
					& \textbf{80.07}     & \textbf{88.47}     & \textbf{59.31}           
					& \textbf{81.73}     & \textbf{90.98}     & \textbf{67.29}\\  \cline{1-17}
					\bottomrule
				\end{tabular}
			}
		\end{table}
	}
	
	\subsubsection{Implementation Details.} 
	
	For both datasets and all comparison methods, we utilized ResNet-18 \cite{He_2016_CVPR} pre-trained on ImageNet\cite{deng2009imagenet} as the backbone, the Adam optimizer with a learning rate of 3e-5, $L_2$ regularization on model parameters, and a batch size of 32. In the federated setting, we configured each client to conduct one local training round before each communication round. The warm-up round $t_1$ was set at 50, and the total number of rounds was 500. The parameters $L$ and $R$ were set to 0.3 and 0.7, respectively, with $T_0$ and $T_1$ fixed at 0.5\% and 1\%. All clients participate in aggregation during each communication round. More details of comparative methods can be found in appendix.
	
	\subsection{Comparison with State-of-the-art Methods}
	
	FedAvg \cite{mcmahan2017communication} and its variant coupled by a partial loss to include only active classes (denoted as FedAVG-PL) are selected as baseline approaches. By regarding missing sample labels as absent class labels, FSSL is extendable to such a setting. Thus, SOTA FSSL methods are introduced for comparison, including RSCFed (CVPR'22) \cite{liang2022rscfed} using a sub-consensus model and local distillation for unknown classes, FedFixMatch by extending FixMatch (NeuIPS'20) \cite{sohn2020fixmatch} with consistency regularization through weak and strong augmentation to FedAvg, FedIRM (MICCAI'21) \cite{liu2021federated} exploring relational class knowledge and aligning clients, and CBAFed (CVPR'23) \cite{li2023class} implementing pseudo-labeling with dynamic thresholds. Our scenario can also be regarded as a setting where all positive samples with unlabeled classes are mislabeled as negative due to noise, addressed by FNNL. SOTA FNNL methods are introduced, including FedLSR (CIKM'22) \cite{jiang2022towards} via self-regularization during local training and FedNoRo (IJCAI'23) \cite{wu2023fednoro} detecting noisy clients with Gaussian mixture model of which only the second stage is trained as all clients are noisy in our setting.
	
	Quantitative comparison measured by balanced accuracy (BACC, for multi-class) and AUC along with mAP (for multi-label) are summarized in Tables \ref{compare ich} and \ref{compare chest}. With increasing class absences, existing FL approaches struggle particularly with less prevalent ``cool classes''. Specifically, FedAvg and FedLSR fail completely with only one class labeled, defaulting to classify all samples as healthy. Comparatively, FedMLP leverages class prototypes to extract knowledge from each client, consistently achieving top or near-top results on two medical image datasets and maintaining strong performance across varying levels of label absence.
	
	\subsection{Ablation Study}
	
	Component-wise ablation study of FedMLP on \textit{ICH} and \textit{ChestXray14} is summarized in Table \ref{abalation}. Each client was configured to focus on a single category, highlighting the contribution of each component under conditions of extreme label scarcity. In such scenarios, FedAvg struggles to detect positive cases, resulting in a decline in BACC down to 50\%. MLD, utilizing class prototypes, empowers clients to access information on unlabeled classes, mitigating the knowledge loss typically incurred during model aggregation. Integrating the remaining three components further improves model efficacy across all evaluation metrics.
	
	{\renewcommand{\arraystretch}{1.1}
		\begin{table}[tbp]
			\centering
			\caption{Quantitative comparison on \textit{ChestXray14} under four levels of partial annotation settings. The results (\%) based on the last-epoch model on the testing set are reported. The best and second-best results are marked in bold and underlined.}\label{compare chest}
			\resizebox{\linewidth}{!}{
				\begin{tabular}{c|c|ccc|ccc|ccc|ccc|ccc}
					\toprule
					\hline
					\multirow{2}{*}{Type} & \multirow{2}{*}{Method}  &  \multicolumn{3}{c|}{Missing 1 class}  &\multicolumn{3}{c|}{Missing 3 classes}        & \multicolumn{3}{c|}{Missing 5 classes}    & \multicolumn{3}{c|}{Missing 7 classes}  & \multicolumn{3}{c}{Average}           
					\\ \cline{3-17} 
					&           & BACC       &AUC        &\multicolumn{1}{c|}{mAP}            
					& BACC       &AUC        &\multicolumn{1}{c|}{mAP}
					& BACC       &AUC        &\multicolumn{1}{c|}{mAP}
					& BACC       &AUC        &\multicolumn{1}{c|}{mAP}
					& BACC       &AUC        &mAP
					\\ \hline
					\multirow{2}{*}{Base} 
					&FedAvg       & 63.27     & 76.46     & 26.89          
					& 57.21     & 75.45     & 23.31      & 51.98     & 70.21     & 18.65      
					& 50.00     & 66.64     & 12.16           
					& 55.62     & 72.19     & 20.25\\  \cline{2-17}
					& FedAvg-PL      & 63.88     & 76.02     & 27.37     & 60.14     & 73.28     & 21.38      & 56.87    & 69.33     & 19.35      
					& 53.30     & 64.80     & 11.49           
					& 58.55     & 70.86     & 19.90  \\  \cline{1-17}
					\multirow{4}{*}{FSSL} 
					&RSCFed        & 65.07     & 77.08     & 26.61          
					& 63.30     & 76.37     & \underline{26.41}      & 58.87     & 70.81     & 23.96      
					& 50.49     & 65.92     & 14.70           
					& 59.43     & 72.55     & 22.92\\  \cline{2-17}
					&FedFixMatch       & 62.90     & 77.94     & 27.70          
					& 58.07     & 75.48     & 25.68      & 58.02     & 69.53     & 19.65     
					& 54.52     & 65.59     & 11.42
					& 58.38     & 72.14     & 21.11\\  \cline{2-17}
					&FedIRM       & 61.62     & 76.03     & 26.99          
					& 57.84     & 73.91     & 23.67      & 54.68     & 67.79     & 20.98      
					& 50.01     & 63.20     & 13.44
					& 56.04     & 70.23     & 21.27\\  \cline{2-17}
					& CBAFed      & \underline{68.77}     &77.33      &26.95      & 62.98      &74.28      &24.02       &52.97     &68.26      &19.67        
					& 50.00    & 52.25     & 7.53                
					& 58.68    & 68.03     & 19.54\\  \cline{1-17}
					\multirow{2}{*}{FNL} 
					&FedLSR       & 62.59     & 76.95     & \underline{28.25}         
					& 55.16     & 74.88     & 25.25      & 52.94     & 70.52     & 19.65      
					& 50.00     & 50.00     & 6.79
					& 55.17     & 68.09     & 19.99\\  \cline{2-17}
					& FedNoRo      & \textbf{69.65}     & \textbf{78.99}     & 27.96 & \textbf{69.27}     & \textbf{78.73}     & 26.26      & \underline{67.68}     & \underline{74.75}     & \underline{25.97} 
					& \underline{66.54}     & \underline{74.23}     & \underline{20.31}       
					& \underline{68.29}     & \underline{76.68}     & \underline{25.13}\\  \cline{1-17}
					\multirow{1}{*}{Ours}
					& FedMLP       &68.55      &\underline{78.44}      &\textbf{28.87}  & \underline{68.37}    & \underline{78.51}     & \textbf{28.01}     & \textbf{68.62}      & \textbf{76.02}    & \textbf{26.63}     & \textbf{70.08}     & \textbf{75.76}      
					& \textbf{22.48}    & \textbf{68.91}     & \textbf{77.18}     & \textbf{26.50}           \\  \cline{1-17}
					\bottomrule
				\end{tabular}
			}
		\end{table}
	}
	
	\begin{table}[tbp]
		\centering
		\caption{Component-wise ablation study.}\label{abalation}
		\resizebox{0.7\linewidth}{!}{
            \renewcommand{\arraystretch}{0.92}
			\begin{tabular}{ccccc|ccc|ccc}
				\toprule
				\hline
				\multirow{2}{*}{FedAvg}  & \multirow{2}{*}{MLD} & \multirow{2}{*}{WPC} & \multirow{2}{*}{CR} & \multirow{2}{*}{ST} 
				& \multicolumn{3}{c|}{\textit{ICH}}       & \multicolumn{3}{c}{\textit{ChestXray14}}         \\ \cline{6-11} 
				&                       &                        &                        &                        
				& BACC                       & AUC           & mAP       
				& BACC                        & AUC           & mAP            \\ \hline
				\checkmark                       &                       &                        &                        &
				& 50.00                    & 76.32          & 32.62
				& 50.00                   & 66.64          & 12.16  \\
				\checkmark                       & \checkmark                     &                      &                        &
				& 73.41                    & 82.32          & 46.74
				& 62.82                    & 69.80          & 13.75  \\
				\checkmark                       & \checkmark                & \checkmark                     &           &       
				& 78.26                   & 85.98          & 52.10
				& 66.63                    & 72.59          & 16.37  \\
				\checkmark                       & \checkmark                & \checkmark            & \checkmark             &        & 79.73                   & 88.44          & 58.81          
				& 68.85                    & 75.06          & 19.56  \\
				\checkmark                       & \checkmark                & \checkmark            & \checkmark     & \checkmark     & \textbf{80.07}         & \textbf{88.47}     & \textbf{59.31}
				& \textbf{70.08}  & \textbf{75.76} & \textbf{22.48}\\ \hline
				\bottomrule
			\end{tabular}
		}
	\end{table}
	
	\section{Conclusion}
	In this paper, we propose a novel federated multi-label partial label setting and identify the catastrophic forgetting problem of "cool classes" with the imbalance of class attention. Our method FedMLP, which combines class prototype generation and global knowledge regularization, enables the majority of clients to leverage their local unlabeled data to enhance limited labeled information. Experimental results demonstrate its superior performance and robustness to high missing rates on two real-world medical datasets. We believe this study is helpful in building real-world FL systems under complicated data/task heterogeneity.
    \\
    \\
    \textbf{Acknowledgement.} This work was supported in part by the National Natural Science Foundation of China under Grants 62202179 and 62271220, and in part by the Natural Science Foundation of Hubei Province of China under Grant 2022CFB585. The computation is supported by the HPC Platform of HUST.
	
	\bibliographystyle{splncs04}
	\bibliography{references}

\begin{thebibliography}{10}
\providecommand{\url}[1]{\texttt{#1}}
\providecommand{\urlprefix}{URL }
\providecommand{\doi}[1]{https://doi.org/#1}

\bibitem{bucak2011multi}
Bucak, S.S., Jin, R., Jain, A.K.: Multi-label learning with incomplete class
  assignments. In: CVPR. pp. 2801--2808. IEEE (2011)

\bibitem{chen2013fast}
Chen, M., Zheng, A., Weinberger, K.: Fast image tagging. In: ICML. pp.
  1274--1282. PMLR (2013)

\bibitem{cho2023local}
Cho, Y.J., Joshi, G., Dimitriadis, D.: Local or global: Selective knowledge
  assimilation for federated learning with limited labels. In: CVPR. pp.
  17087--17096 (2023)

\bibitem{deng2009imagenet}
Deng, J., Dong, W., Socher, R., Li, L.J., Li, K., Fei-Fei, L.: Imagenet: A
  large-scale hierarchical image database. In: CVPR. pp. 248--255. Ieee (2009)

\bibitem{flanders2020construction}
Flanders, A.E., Prevedello, L.M., Shih, G., Halabi, S.S., Kalpathy-Cramer, J.,
  Ball, R., Mongan, J.T., Stein, A., Kitamura, F.C., Lungren, M.P., et~al.:
  Construction of a machine learning dataset through collaboration: the rsna
  2019 brain ct hemorrhage challenge. Radiology: Artificial Intelligence
  \textbf{2}(3),  e190211 (2020)

\bibitem{He_2016_CVPR}
He, K., Zhang, X., Ren, S., Sun, J.: Deep residual learning for image
  recognition. In: CVPR (2016)

\bibitem{FCCL_CVPR22}
Huang, W., Ye, M., Du, B.: Learn from others and be yourself in heterogeneous
  federated learning. In: CVPR (2022)

\bibitem{FCCLPlus_TPAMI23}
Huang, W., Ye, M., Shi, Z., Du, B.: Generalizable heterogeneous federated
  cross-correlation and instance similarity learning. IEEE TPAMI  (2023)

\bibitem{FPL_CVPR23}
Huang, W., Ye, M., Shi, Z., Li, H., Du, B.: Rethinking federated learning with
  domain shift: A prototype view. In: CVPR (2023)

\bibitem{FLSurveyandBenchmarkforGenRobFair_arXiv23}
Huang, W., Ye, M., Shi, Z., Wan, G., Li, H., Du, B., Yang, Q.: A federated
  learning for generalization, robustness, fairness: A survey and benchmark.
  arXiv  (2023)

\bibitem{jiang2022dynamic}
Jiang, M., Yang, H., Li, X., Liu, Q., Heng, P.A., Dou, Q.: Dynamic bank
  learning for semi-supervised federated image diagnosis with class imbalance.
  In: MICCAI. pp. 196--206 (2022)

\bibitem{jiang2022towards}
Jiang, X., Sun, S., Wang, Y., Liu, M.: Towards federated learning against noisy
  labels via local self-regularization. In: CIKM. pp. 862--873 (2022)

\bibitem{li2023class}
Li, M., Li, Q., Wang, Y.: Class balanced adaptive pseudo labeling for federated
  semi-supervised learning. In: CVPR. pp. 16292--16301 (2023)

\bibitem{liang2022rscfed}
Liang, X., Lin, Y., Fu, H., Zhu, L., Li, X.: Rscfed: Random sampling consensus
  federated semi-supervised learning. In: CVPR. pp. 10154--10163 (2022)

\bibitem{liu2021federated}
Liu, Q., Yang, H., Dou, Q., Heng, P.A.: Federated semi-supervised medical image
  classification via inter-client relation matching. In: MICCAI. pp. 325--335
  (2021)

\bibitem{mcmahan2017communication}
McMahan, B., Moore, E., Ramage, D., Hampson, S., y~Arcas, B.A.:
  Communication-efficient learning of deep networks from decentralized data.
  In: AISTATS. pp. 1273--1282 (2017)

\bibitem{menon2020long}
Menon, A.K., Jayasumana, S., Rawat, A.S., Jain, H., Veit, A., Kumar, S.:
  Long-tail learning via logit adjustment. In: ICLR (2021)

\bibitem{sohn2020fixmatch}
Sohn, K., Berthelot, D., Carlini, N., Zhang, Z., Zhang, H., Raffel, C.A.,
  Cubuk, E.D., Kurakin, A., Li, C.L.: Fixmatch: Simplifying semi-supervised
  learning with consistency and confidence. In: NeurIPS. vol.~33, pp. 596--608
  (2020)

\bibitem{sun2010multi}
Sun, Y.Y., Zhang, Y., Zhou, Z.H.: Multi-label learning with weak label. In:
  AAAI. vol.~24, pp. 593--598 (2010)

\bibitem{wang2014binary}
Wang, Q., Shen, B., Wang, S., Li, L., Si, L.: Binary codes embedding for fast
  image tagging with incomplete labels. In: ECCV. pp. 425--439. Springer (2014)

\bibitem{wang2017chestx}
Wang, X., Peng, Y., Lu, L., Lu, Z., Bagheri, M., Summers, R.M.: Chestx-ray8:
  Hospital-scale chest x-ray database and benchmarks on weakly-supervised
  classification and localization of common thorax diseases. In: CVPR. pp.
  2097--2106 (2017)

\bibitem{wu2023learning}
Wu, C., Li, Z., Wang, F., Wu, C.: Learning cautiously in federated learning
  with noisy and heterogeneous clients. arXiv preprint arXiv:2304.02892  (2023)

\bibitem{FedISM}
Wu, N., Kuang, Z., Yan, Z., Yu, L.: From optimization to generalization: Fair
  federated learning against quality shift via inter-client sharpness matching.
  In: IJCAI (2024)

\bibitem{FedA3I}
Wu, N., Sun, Z., Yan, Z., Yu, L.: Feda3i: Annotation quality-aware aggregation
  for federated medical image segmentation against heterogeneous annotation
  noise. In: AAAI (2024)

\bibitem{wu2023fednoro}
Wu, N., Yu, L., Jiang, X., Cheng, K.T., Yan, Z.: Fednoro: Towards noise-robust
  federated learning by addressing class imbalance and label noise
  heterogeneity. IJCAI  (2023)

\bibitem{wu2023federated}
Wu, N., Yu, L., Yang, X., Cheng, K.T., Yan, Z.: Fediic: Towards robust
  federated learning for class-imbalanced medical image classification. In:
  MICCAI (2023)

\bibitem{xu2022fedcorr}
Xu, J., Chen, Z., Quek, T.Q., Chong, K.F.E.: Fedcorr: Multi-stage federated
  learning for label noise correction. In: CVPR. pp. 10184--10193 (2022)

\end{thebibliography}

\end{document}


\chapter*{Supplementary Material}

\begin{table}[htbp]
	\centering
	
	\centering
	\caption{Implementation details of some comparison methods.}\label{abalation2}
	\begin{tabular}{l|c}
		\toprule
		\hline
		\multirow{1}{*}{Methods}  & \multirow{1}{*}{Details}          \\ \cline{1-2} 
		RSCFed		& $K$=3, $M$=3 for \textit{ICH} and $K$=6, $M$=10 for \textit{ChestXray14} \\ \cline{1-2} 
		FixMatch		& $\tau$=0.8 \\ \cline{1-2} 
		FedIRM		& $T$=20, $\tau$=2.0, $\omega$=30, $ema$=0.99 \\ \cline{1-2} 
		CBAFed		& $P$=50, $\alpha_1$=0.8, $\alpha_2$=0.5, $J$=1, $T$=500, $\tau$=0.6 \\ \cline{1-2} 
		FedLSR		& $\lambda\sim Beta(1,1)$, $\gamma$=$Min$[0.4$\cdot$$\frac{t}{40}$, 0.4] \\ \cline{1-2} 
		FedNoRo		& $\lambda_{max}$=0.8 \\ \cline{1-2} 
		
		\bottomrule
	\end{tabular}
	
\end{table}

\begin{figure}[h]
	\centering
	\includegraphics[width=1.0\textwidth]{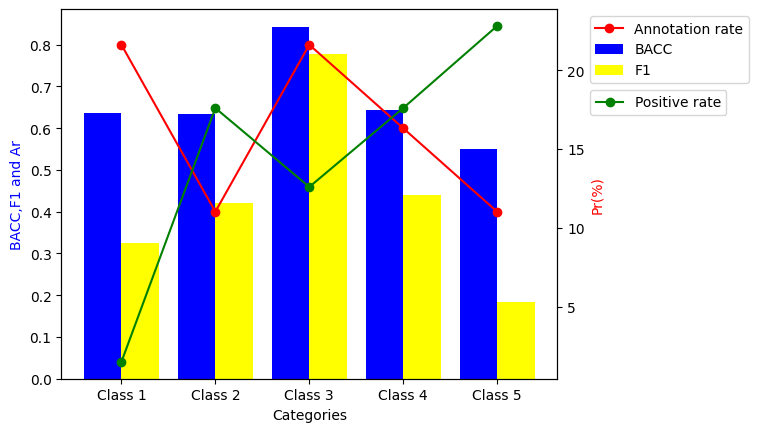}
	\caption{Results of FedAvg on \textit{ICH} where cool classes with lower annotation rates
		facing performance degradation. The recognition ability of class 1 is also weak due to
		the influence of intra-class imbalance.}
	\label{output}
\end{figure}

\begin{algorithm}[htbp]
	\caption{FedMLP.}
	\label{alg:NG}
	\textbf{Input}: Initialized global model $\theta^1_G$; dataset $D_k$ in client $k$ amount $N_k$, $k \in [K]$; labeled class set $AC_k$ and unlabeled class set $NC_k$ in client $k$; hyper-parameter $L$, $R$, $T_0$ and $T_1$; Warm-up rounds $t_1$; total communication rounds $T$; local training epoch $E$.
	\begin{algorithmic}[1] 
		\item\textcolor{blue}{$\triangleright$ \textit{Stage1: Warm-up}} \vfill 
		\FOR{$t$ $\leftarrow$ 1 to $t_1$}
		\STATE \textbf{In local clients}
		\FOR{$k$ $\leftarrow$ 1 to $K$}
		\STATE $\theta^t_k$ $\leftarrow$ $\theta^t_G$
		\FOR{$e$ $\leftarrow$ 1 to $E$}
		\STATE Augment for each sample in $D_k$ as $D^{Aug_1}_k$ and $D^{Aug_2}_k$
		\STATE $\theta^t_k\leftarrow$ update by Eq. \textcolor{red}{4} with the local dataset $D^{Aug_1}_k \cup D^{Aug_2}_k$
		\ENDFOR
		\IF{$t$ = $t_1$}
		\STATE Local Calculation()
		\ENDIF
		\ENDFOR
		\STATE \textbf{In central server}
		\IF{$t$ = $t_1$}
		\STATE Global Aggregation($\theta^t_k$, $P^{k,c}_0$, $P^{k,c}_1$, $d^k_c$)
		\ELSE 
		\STATE $\theta^{t+1}_G \leftarrow \sum_{k=1}^{K} \frac{N_k}{\sum_{i=1}^{K} N_i} \theta^t_k$
		\ENDIF
		\ENDFOR
		
		\item\textcolor{blue}{$\triangleright$ \textit{Stage2: Missing Label Detection}} \vfill
		\FOR{$t$ $\leftarrow$ $t_1 + 1$ to $T$}
		\STATE \textbf{In local clients}
		\FOR{$k$ $\leftarrow$ 1 to $K$}
		\STATE Download $\tau_0$, $\tau_1$, $\theta^t_G$, $P^c_0$, $P^c_1$ and $d^G_c$
		\STATE select samples and categories with pseudo-labels use Eq. \textcolor{red}{7} $\sim$ \textcolor{red}{9}
		\FOR{$e$ $\leftarrow$ 1 to $E$}
		\STATE $\theta^t_k\leftarrow$ updated by Eq. \textcolor{red}{4} for hard labels and MSE for soft labels 
		\ENDFOR
		\STATE Local Calculation()
		\ENDFOR
		\STATE \textbf{In central server}
		\STATE Global Aggregation($\theta^t_k$, $P^{k,c}_0$, $P^{k,c}_1$, $d^k_c$)
		\ENDFOR
		\STATE 
		\STATE
		
		\textbf{Global Aggregation($\theta^t_k$, $P^{k,c}_0$, $P^{k,c}_1$, $d^k_c$)}
		\STATE \quad $\theta^{t+1}_G \leftarrow \sum_{k=1}^{K} \frac{N_k}{\sum_{i=1}^{K} N_i} \theta^t_k$
		\STATE \quad Calculate $P^c_0$, $P^c_1$ and $d^G_c$ use Eq. \textcolor{red}{6} and \textcolor{red}{10}
		\STATE \quad Calculate $\tau_0$ and $\tau_1$
		\STATE \textbf{Output}: $\theta^{t+1}_G$, $P^c_0$, $P^c_1$, $d^G_c$, $\tau_0$ and $\tau_1$.  
		\STATE
		\STATE
		
		\textbf{Local Calculation()}
		\STATE \quad Calculate $P^{k,c}_0$ and $P^{k,c}_1$ for $c \in AC_k$ use Eq. \textcolor{red}{5}
		\STATE \quad Calculate $d^k_c$ for $c \in AC_k$ use Eq. \textcolor{red}{10}
		\STATE \textbf{Output}: $P^{k,c}_0$, $P^{k,c}_1$ and $d^k_c$. 
	\end{algorithmic}
	\textbf{Output}: Global model $\theta^{T+1}_G$. 
\end{algorithm}